\documentclass[graybox]{svmult}
\usepackage{mathptmx}       %
\usepackage{helvet}         %
\usepackage{courier}        %
\usepackage{type1cm}        %
\usepackage{makeidx}         %
\usepackage{graphicx}        %
\usepackage{multicol}        %
\usepackage[bottom]{footmisc}%

\usepackage{hyperref}

\makeindex             %

\usepackage{titlesec}
\titlespacing\section{0pt}{4pt plus 4pt minus 2pt}{2pt plus 0pt minus 2pt}
\titlespacing\subsection{0pt}{4pt plus 4pt minus 2pt}{2pt plus 0pt minus 2pt}
\titlespacing\subsubsection{0pt}{4pt plus 4pt minus 2pt}{2pt plus 0pt minus 2pt}
     
\pdfoutput=1

\usepackage[numbers]{natbib}
\usepackage{multicol}

\usepackage[]{graphicx}
\usepackage{enumerate}
\usepackage{subfigure}
\usepackage{amssymb}
\usepackage[cmex10]{amsmath}

\graphicspath{{./PLOTS/}}

\usepackage{algorithmic}
\usepackage{algorithm}

\usepackage{times}

\usepackage[numbers]{natbib}
\usepackage{multicol}
\usepackage{enumerate}
\usepackage{subfigure}
\usepackage{graphicx}
\usepackage{amsmath,amssymb}%
\usepackage{verbatim}
\usepackage{bm}
\usepackage{algorithmic}
\usepackage{algorithm}

\newcommand{\FMT}{FMT$^*\,$}
\newcommand{\real}{{\mathbb{R}}}
\newcommand{\reals}{\real}

\newcommand{\bs}{\boldsymbol}
\newcommand{\var}{\ensuremath{\operatorname{Var}}}
\newcommand{\Xfree}{\mbox{$\mathcal{X}_{\text{free}}$}}
\newcommand{\Xgoal}{\mbox{$\mathcal{X}_{\text{goal}}$}}
\newcommand{\Xobs}{\mbox{$\mathcal{X}_{\text{obs}}$}}
\newcommand{\xinit}{\mbox{$\bs{x}_0$}}
\renewcommand{\prob}[1]{\mathbb{P}\left( #1 \right) }
\DeclareMathOperator*{\argmin}{arg\,min}

\newcommand{\x}{\mathbf{x}}
\renewcommand{\u}{\mathbf{u}}
\newcommand{\y}{\mathbf{y}}
\newcommand{\dx}{\delta\mathbf{x}}
\newcommand{\du}{\delta\mathbf{u}}
\newcommand{\dy}{\delta\mathbf{y}}
\newcommand{\xn}{\mathbf{x}^{\text{nom}}}
\newcommand{\un}{\mathbf{u}^{\text{nom}}}
\newcommand{\yn}{\mathbf{y}^{\text{nom}}}

\newcommand{\z}{\mathbf{z}}

\renewcommand{\v}{\mathbf{v}}
\newcommand{\n}{\mathbf{n}}
\newcommand{\w}{\mathbf{w}}
\newcommand{\dt}{{\Delta t}}
\newcommand{\dm}{{\Delta \mu}}

\makeatletter
\g@addto@macro\normalsize{%
  \setlength\abovedisplayskip{1pt}
  \setlength\belowdisplayskip{1pt}
  \setlength\abovedisplayshortskip{1pt}
  \setlength\belowdisplayshortskip{1pt}
}

\begin{document}
\title*{Monte Carlo Motion Planning for Robot Trajectory Optimization Under Uncertainty}
\titlerunning{Monte Carlo Motion Planning} 
\author{Lucas Janson$^\dagger$, Edward Schmerling$^\dagger$, and Marco Pavone}
\authorrunning{Lucas Janson, Edward Schmerling, and Marco Pavone}
\institute{Lucas Janson \at Department of Statistics, Stanford University, Stanford, CA 94305, \email{ljanson@stanford.edu}
\and Edward Schmerling \at Institute for Computational and
Mathematical Engineering, Stanford University, Stanford, CA 94305, \email{schmrlng@stanford.edu}
\and Marco Pavone \at Department of Aeronautics and Astronautics, Stanford University, Stanford, CA 94305, \email{pavone@stanford.edu}
\and  $^\dagger$ Lucas Janson and Edward Schmerling contributed equally to this work.\\\\This work was supported  by NASA under the Space Technology Research Grants Program, Grant NNX12AQ43G. Lucas Janson was partially supported by NIH training grant T32GM096982.}

\maketitle

\vspace{-.25cm}
\begin{abstract}
{This article presents a novel approach, named MCMP (Monte Carlo Motion
Planning), to the problem of motion planning under uncertainty, i.e.,
to the problem of computing a low-cost path that fulfills
probabilistic collision avoidance constraints. MCMP estimates
the collision probability (CP) of a given path by sampling  via Monte
Carlo the execution of a reference tracking controller (in this
paper we consider LQG). The key algorithmic contribution of this paper is the design of
statistical variance-reduction techniques, namely control variates
and importance sampling, to make such a sampling procedure amenable to
real-time implementation. MCMP applies this CP estimation procedure to motion planning by iteratively  (i) computing an
(approximately) optimal path for the deterministic version of the
problem (here, using the \FMT algorithm), (ii) computing the CP of
this path, and (iii) inflating or deflating the obstacles by a common
factor depending on whether the CP is higher or lower than a target
value. The advantages of MCMP are threefold: (i) asymptotic
correctness of CP estimation, as opposed to most current
approximations, which, as shown in this paper, can be off by large
multiples and hinder the computation of feasible plans; (ii) speed and
parallelizability, and (iii) generality, i.e., the approach is
applicable to virtually any planning problem provided that a path
tracking controller and a notion of distance to obstacles in the
configuration space are available. Numerical results
illustrate the correctness (in terms of feasibility), efficiency (in
terms of path cost), and computational speed of MCMP.}

\end{abstract}

\section{Introduction}
Robotic motion planning is the problem of computing a path
that connects an initial and a terminal robot state while avoiding collisions with
obstacles and optimizing an objective function \cite{SML:06}. Despite
the fact that finding a feasible, let alone optimal, solution to a
motion planning problem is difficult (even the most basic versions of
the problem are already PSPACE-hard \cite{JHR:79, SML:06}), in the
past three decades key breakthroughs have made the solution to this
problem largely practical (see \cite{SML:06} and references therein
for a comprehensive historical account). Most works in the literature,
however, focus on a deterministic setup where the state of a robot is
perfectly known and its actions lead to a deterministic, unique
outcome. While this is usually an excellent approximation for robots
operating in highly structured environments (e.g., manipulators in an
assembly line), it falls short in unstructured settings, e.g., for
ground or aerial field robots or surgical robotic systems
\cite{SML:11b}. In such cases, motion uncertainty, sensing
uncertainty, and environment uncertainty may dramatically alter the
safety and quality of a path computed via deterministic techniques
(i.e., neglecting uncertainty). Hence, accounting for uncertainty in
the planning process is regarded as an essential step for the
deployment of robotic systems ``outside the factory
  floor'' \cite{SML:11b}. In this paper we introduce Monte Carlo Motion Planning, a novel approach to planning under uncertainty that is accurate, fast, and general.

{\em Related work:} Conceptually, to enable a robot to plan its motion under uncertainty,
one needs to design a strategy for a decision maker. In this regard, robotic motion planning
can be formalized as a partially observable Markov decision process
(POMDP) \cite{LPK-MLL-ARC:98}, where the key idea is to assume that
the state evolves according to a controlled Markov chain,
the state is only partially observed, and one seeks to design a
 \emph{control policy} that maps state probability distributions to actions. However, despite the theoretical
\cite{LPK-MLL-ARC:98} and practical \cite{HK-DH-WSL:08} successes of
the POMDP theory, the online computation of a control policy for robotic applications is extremely computationally intensive, and possibly even unnecessary  as after a short time horizon the environment map may have changed \cite{SML:06}.
The alternative and widely adopted approach is then to restrict the optimization process to \emph{open-loop} trajectories, which involves the much simpler task of computing  a \emph{control sequence} (as opposed to a control policy), and recompute the reference trajectory in a receding horizon fashion (e.g., every few seconds). This is the approach we consider in this paper.

To select open-loop trajectories, a large number of works cast the problem into a
chance-constrained optimization problem
\cite{LB-MO-AB-BCW:10}, where under the assumption of linear dynamics and
convex obstacles, an open-loop control sequence is computed as the
solution to a mixed-integer linear program. The works in
\cite{MPV-CJT:11, MO-BCW-LB:13} extend this approach to an optimization over
the larger class of affine output feedback controllers, comprising a
nominal control input and an error feedback term. These works,
however, require an explicit characterization of the obstacle space
(in the configuration space), which is oftentimes unavailable
\cite[Chapter 5]{SML:06}. This has prompted a number of
researchers to extend the sampling-based motion planning paradigm to the problem of 
planning under uncertainty (in the sampling-based paradigm, an \emph{explicit} construction of the configuration space is avoided and the configuration space is probabilistically ``probed" with a sampling scheme \cite{SML:06}).  A common approach is to forgo path
optimization and recast the problem as an \emph{unconstrained}
planning problem where the path collision probability (CP) is
minimized. For example, the approach of LQG-MP \cite{JVDB-PA-KG:11} is
to approximate a path CP by combining pointwise CPs as if they were
independent, running the rapidly-exploring random trees (RRT)
\cite{SML-JJK:01} algorithm multiple times, and then selecting the path
with minimum (approximate) path CP. The pointwise CPs
are computed within the model that  a reference tracking controller is
employed to track a nominal open-loop path. This is closely related
to model predictive control (MPC) with closed-loop prediction
\cite{FO-CJ-MM:08} and  leads to a less conservative collision probability estimate than if the nominal control was executed without feedback. A
similar approach is used in \cite{SW-LGT-JVDB-RA:13}, where the
authors employ a truncation method \cite{SP-JVDB-RA:12} to improve
the accuracy of path CP computation.

The interplay between minding collision probability while simultaneously
optimizing a path planning cost objective function is
considered in \cite{WS-SP-RA:15}, although still with an
approximation to the path CP. There, cost optimization is considered
over the set of path plans satisfying a lower bound on success probability.
The inclusion of path CP as a
constraint is also considered in \cite{BL-MK-JPH:10}, where the authors
propose CC-RRT, an RRT-based algorithm that approximates path CP via
Boole's bound. CC-RRT has been extended to include dynamic
obstacles via Bayesian nonparametric models
\cite{GSA-BDL-JMJ-NR-JPH:13}, tailored to the control of unmanned
aerial vehicles \cite{MK-IP:13} and  parafoils \cite{BDL-IS-JPH:13},
and combined with the RRT* asymptotically optimal version of RRT \cite{BDL-SK-JPH:13}.

{\em Contributions:} In this paper we present an algorithm for robot planning under uncertainty that returns high quality solutions (in terms of a general planning objective) which satisfy a specified constraint on collision probability, or safety tolerance. The motivation of this work is that all of the aforementioned approaches approximate path CP in ways that can be quite inaccurate, thus potentially drastically mischaracterizing the feasible domain of path optimization. In particular, we show (see Figure~\ref{fig:CP_results}) that those approximations can be off by \emph{many}
multiples in simple examples. To address this problem, our first contribution is to design a variance-reduced (that is, quickly-converging) Monte Carlo (MC) probability estimation algorithm for CP computation.  This algorithm  estimates the collision probability of a given trajectory by sampling many
realizations of a reference-tracking controller, modeling the effort of a robot to follow a reference path. In particular, in
this paper, we assume a Linear-Quadratic Gaussian (LQG) tracking
controller, similar to LQG-MP \cite{JVDB-PA-KG:11} and
MPC with closed-loop prediction
\cite{FO-CJ-MM:08}.
Our algorithm does not suffer the
inaccuracies of the approximations mentioned earlier, and indeed
provides the exact path CP given enough time (in contrast to current
approaches). Most importantly, our variance-reduction scheme, which
combines and tailors control variate and importance
sampling techniques in an original fashion to the problem at hand, enables the
computation of very accurate estimates in a way compatible with real-time operations.
This holds even when working with very small CPs, a regime in which a straightforward Monte Carlo
method would require great computational expense to arrive at accurate estimates.
Another key advantage of our algorithm is that it comes with an estimate
of its variance, so that we have a measure of accuracy,
unlike the aforementioned approximations. It is also trivially
parallelizable and has the potential to be extended to very general
controllers and uncertainty models.

Our estimation algorithm 
enables a novel approach to planning under uncertainty, which we call
Monte Carlo Motion Planning (MCMP)---our second contribution. MCMP proceeds by performing bisection
search over CP and obstacle inflation, at each step solving a
deterministic version of the problem with inflated
obstacles. To demonstrate the performance of MCMP, we present simulation results that
illustrate the correctness (in terms of feasibility), efficiency (in
terms of path cost), and computational speed of MCMP. From a
conceptual standpoint, MCMP can be viewed as a planning analogue to
MC approaches for robot localization \cite{DF-WB-FD-ST:01}.

{\em Organization:} This paper is structured as follows. Section
\ref{sec:background} reviews some background on MC variance
reduction. Section \ref{sec:problem} formally defines the problem we
consider in this paper. Section~\ref{sec:pathCP} elucidates the
shortcomings of previous path CP approximation schemes. Section
\ref{sec:main} presents variance-reduction techniques for fast MC
computation of path CP. Section \ref{sec:algo} presents the
overall MCMP approach. Section \ref{sec:simulations} presents results
from numerical experiments supporting our statements. Finally, in
Section \ref{sec:conclusion}, we draw some conclusions and discuss
directions for future work.

\section{Background on Monte Carlo Variance Reduction}
\label{sec:background}
The use of Monte Carlo (MC) to estimate the probability of complex events is
well-studied. In this section we will briefly introduce MC and the two
variance reduction techniques that provide the basis for our main
result in Section~\ref{sec:main}. For more detail and other topics on Monte
Carlo, the reader is referred to the excellent unpublished text
\cite{ABO:13}, from which the material of this section is taken.

\subsection{Simple Monte Carlo}
In its most general form, MC is a way of estimating the expectation of a function of a
random variable by drawing many independent and identically
distributed (i.i.d.) samples of that random variable, and averaging their
function values. Explicitly, consider a random variable $\bs{X} \in
\mathbb{R}^n$ and a bounded function $f:\mathbb{R}^n \rightarrow
\mathbb{R}$. For a sequence of $m$ i.i.d. realizations
of $\bs{X}$, $\{\bs{X}^{(i)}\}_{i=1}^m$, the central limit theorem gives,
\begin{equation}
\label{simple}
\sqrt{m}\left(\frac{1}{m}\sum_{i=1}^mf\left(\bs{X}^{(i)}\right) -
  \mathbb{E}\left[f\left(\bs{X}\right)\right]\right) \stackrel{\mathcal{D}}{\longrightarrow} N\left(0, \tau^2\right),
\end{equation}
as $m \rightarrow \infty$, where $\stackrel{\mathcal{D}}{\longrightarrow}$
denotes convergence in distribution, and $N(0,\tau^2)$ refers to the
Gaussian distribution with mean $0$ and variance $\tau^2$. This implies
$\frac{1}{m}\sum_{i=1}^mf\left(\bs{X}^{(i)}\right)
\stackrel{p}{\longrightarrow}
\mathbb{E}\left[f\left(\bs{X}\right)\right]$ as $m \rightarrow \infty$, where
$\stackrel{p}{\longrightarrow}$ denotes convergence in
probability. 

In this paper, $\bs{X}$ will be a random trajectory controlled to follow a
nominal path, and $f$ will be the indicator function that a trajectory
collides with an obstacle; call this collision event $A$. Therefore,
the expectation in Equation~\eqref{simple} is just
$\mathbb{E}[f(\bs{X})] = \mathbb{P}(A)$. Denote this
collision probability by $p$, and define
$\hat{p}_{\text{simple}}:=\frac{1}{m}\sum_{i=1}^mf\left(\bs{X}^{(i)}\right)$. Then $\tau^2$ can be consistently estimated
by the sample variance of the $f(\bs{X}^{(i)})$,
\begin{equation}
\hat{\tau}^2 := \frac{1}{m}\sum_{i=1}^m\left(f\left(\bs{X}^{(i)}\right) - \hat{p}_{\text{simple}}\right)^2 \stackrel{p}{\longrightarrow} \tau^2,
\end{equation}
as $m \rightarrow \infty$. $\hat{V}_{\text{simple}} :=
\hat{\tau}^2/m$ allows us to quantify the uncertainty in the CP estimator
$\hat{p}_{\text{simple}}$ by approximating its variance. The material
from this subsection can be found with more detail in \cite[Chapter 2]{ABO:13}.

\subsection{Control Variates}
\label{sec:cv.background}
To reduce the variance of $\hat{p}_{\text{simple}}$, we can use the
method of control variates (CV). CV requires a function $h:\mathbb{R}^n \rightarrow
\mathbb{R}$ such that $\theta := \mathbb{E}[h(\bs{X})]$ is
known. Then if $h(\bs{X}^{(i)})$ is correlated with $f(\bs{X}^{(i)})$, its
variation around its (known) mean can be used to characterize the
variation of $f(\bs{X}^{(i)})$ around its (unknown) mean, which can
then be subtracted off from $\hat{p}_{\text{simple}}$. Explicitly, given a scaling
parameter value $\beta$, we estimate $p = \mathbb{E}[f(\bs{X})]$ by,
\begin{equation}
\hat{p}_{\beta} := \frac{1}{m}\sum_{i=1}^m(f(\bs{X}^{(i)}) - \beta
h(\bs{X}^{(i)})) + \beta\theta = \hat{p}_{\text{simple}} - \beta(\hat{\theta}-\theta),
\end{equation}
where $\hat{\theta}$ is the sample average of the $h(\bs{X}^{(i)})$. The optimal
(variance-minimizing) choice of $\beta$ can be estimated from the simulated data as
\begin{equation}
\hat{\beta} := \sum_{i=1}^m (f(\bs{X}^{(i)}) - \hat{p}_{\text{simple}})(h(\bs{X}^{(i)})
- \hat{\theta})\Big/ \sum_{i=1}^m(h(\bs{X}^{(i)}) - \hat{\theta})^2.
\end{equation}
We then use the CP estimator $\hat{p}_{\hat{\beta}}$, whose variance can
be estimated by,
\begin{equation}
\hat{V}_{\hat{\beta}} := \frac{1}{m^2}\sum_{i=1}^m(f(\bs{X}^{(i)}) -
\hat{p}_{\hat{\beta}} - \hat{\beta}(h(\bs{X}^{(i)}) - \hat{\theta}))^2.
\end{equation}
The data-dependent choice of $\beta$ introduces a bias in
$\hat{p}_{\hat{\beta}}$ that is
asymptotically (in $m$) negligible compared to its variance,
so we will ignore it here. As $m \rightarrow \infty$,
the variance reduction due to CV can be characterized by $\var(\hat{p}_{\hat{\beta}})/\var(\hat{p}_{\text{simple}}) \rightarrow
1-\rho^2$, where $\rho$ is the correlation between $f(\bs{X})$ and
$h(\bs{X})$. The material from this subsection can be found with more
detail in \cite[Section 8.9]{ABO:13}.

\subsection{Importance Sampling}
When $f(\bs{X})$ is the
indicator function for a rare event $A$, as it is in this paper (we assume
that in most settings, path CP constraints will be small to ensure a
high likelihood of safety), MC variance reduction
is often needed, with importance sampling (IS) a particularly useful
tool. Since $p \ll 1$, we can approximate the coefficient of variation
(ratio of standard deviation to expected value) of the estimator $\hat{p}_{\text{simple}}$ as,
\begin{equation}
\frac{\sqrt{\var(\hat{p}_{\text{simple}})}}{\mathbb{E}[\hat{p}_{\text{simple}}]}
= \frac{\sqrt{p(1-p)/m}}{p} \approx \frac{1}{\sqrt{m}}\frac{1}{\sqrt{p}}.
\end{equation}
This means that in order to get the \emph{relative} uncertainty in
$\hat{p}_{\text{simple}}$ to be small, one needs $m \gg 1/p$ which can
be very large, and this
is simply due to the rarity of observing the event $A$. IS allows us
to sample $\bs{X}$ from a distribution that makes the event $A$ more
common and still get an unbiased estimate of $p$.

Until now we have considered $\bs{X}$ to have some fixed probability
density function (pdf) $P:\mathbb{R}^n \rightarrow
\mathbb{R}_{\ge0}$. Denoting the expectation of $f(\bs{X})$ when
$\bs{X}$ has pdf $P$ by $\mathbb{E}_{P}[f(\bs{X})]$,
then for any pdf $Q$ whose support
contains that of $P$ (that is, $P(\bs{x})>0 \Rightarrow Q(\bs{x})>0$),
\begin{gather*}
\mathbb{E}_{P}[f(\bs{X})] = \int_{\mathbb{R}^n} f(\bs{x}) P(\bs{x})
d\bs{x} = \int_{\mathbb{R}^n} f(\bs{x}) \frac{P(\bs{x})}{Q(\bs{x})}
Q(\bs{x}) d\bs{x} 
= \mathbb{E}_{Q}\left[f(\bs{X}) \frac{P(\bs{X})}{Q(\bs{X})}\right]
\end{gather*}
Therefore, letting $\{\tilde{\bs{X}}^{(i)}\}_{i=1}^m$ be
i.i.d. samples with pdf $Q$, the IS
estimate and associated variance estimate are,
\begin{equation}
\hat{p}_Q := \frac{1}{m}\sum_{i=1}^m \frac{f(\tilde{\bs{X}}^{(i)})
  P(\tilde{\bs{X}}^{(i)})}{Q(\tilde{\bs{X}}^{(i)})},
\qquad \quad \hat{V}_Q := \frac{1}{m^2}\sum_{i=1}^m \left(\frac{f(\tilde{\bs{X}}^{(i)})
  P(\tilde{\bs{X}}^{(i)})}{Q(\tilde{\bs{X}}^{(i)})} - \hat{p}_Q\right)^2.
\end{equation}
If $Q$ can be chosen in such a way that $A$ is common \emph{and} the
likelihood ratio $P(\bs{X})/Q(\bs{X})$ does not have high variance for
$\bs{X}\in A$, then $\var(\hat{p}_Q)$ can be much smaller (orders of
magnitude) than $\var(\hat{p}_{\text{simple}})$ for the same $m$. The material from this subsection can
be found with more detail in \cite[Chapter 9]{ABO:13}.

\subsection{Comments}
CV and IS may be combined into one estimator, summarized in Algorithm~\ref{mc}, the
full mathematical details of which are contained in \cite[Section
9.10]{ABO:13}. Although CV and IS can both be excellent frameworks for variance
reduction in MC, there is no general method for selecting $h$ or $Q$,
and good choices for either one are extremely problem-dependent. Indeed, the
main contribution of this paper is to find, for the important case of linear
dynamics and Gaussian noise, $h$ and $Q$ that make MC
estimation of CPs converge fast enough for real-time planning.

\section{Problem Statement}
\label{sec:problem}

We pose the problem of motion planning under uncertainty with safety tolerance as a constraint separate from the path cost to be optimized. We consider robots described by linear dynamics with control policies derived as LQG controllers tracking nominal trajectories. These nominal trajectories are planned assuming continuous dynamics, but in order to make the computation of path CPs tractable, we assume discretized (zero-order hold) approximate dynamics for the tracking controllers
The full details of the continuous vs. discrete problem formulations are rather standard and due to space limitations are provided in Appendix~\ref{APP:problem} of the extended version of this paper \cite{LJ-ES-MP:15EV}. Briefly here, with $\mathcal{N}(\mu, \Sigma)$ denoting a multivariate Gaussian with mean $\mu$ and covariance matrix $\Sigma$, the system dynamics are given by
\begin{equation}\label{eqn:DLQG}
\x_{t+1} = A\x_t + B\u_t+\v_t, \quad \v_t \sim \mathcal{N}(\mathbf{0},
V), \quad \y_t = C\x_t + \w_t, \quad \w_t \sim \mathcal{N}(\mathbf{0}, W).
\end{equation}
where $\x_t \in \reals^d$ is the state, $\u_t \in \reals^\ell$ is the control input, $\y_t$ is the workspace output, and $\v_t$ and $\w_t$
represent Gaussian process and measurement noise, respectively.
With deviation variables from a nominal trajectory defined as $\dx_t:=\x_t - \xn_t$, $\du_t:=\u_t - \un_t$, and $\dy_t:=\y_t - \yn_t$, for $t = 0,\dots,T$, the discrete LQG controller $\du^{\text{LQG}}_t := L_t \, \widehat \dx_t$, with $L_t$ and $\widehat \dx_t$ denoting the feedback gain matrix and Kalman state estimate respectively, minimizes the tracking cost function
\[
J := \mathbb{E}\left[\dx_T^T F \dx_T + \sum_{t=0}^{T-1} \dx_t^T Q \dx_t + \du_t^T R \du_t \right].
\]
The computation details of $L_t$ and the dynamics of $\widehat \dx_t$ are standard and given in Appendix~\ref{APP:problem} \cite{LJ-ES-MP:15EV}; in the remainder of this paper we use only the notation that the combined state/estimate deviations evolve as multivariate Gaussians $\left[\dx_{t}; \widehat \dx_{t} \right] \sim \mathcal{N}\left(\mu_t, \Sigma_t\right)$ and for suitable definitions of $M_t$ and $N_t$ we may write
\begin{equation}\label{eqn:uncertaintyevol}
\begin{bmatrix} \dx_{t+1}\\ \widehat \dx_{t+1} \end{bmatrix} \sim \mathcal{N}\left(\mu_{t+1} = M_t\mu_t, \Sigma_{t+1} = M_t\Sigma_tM_t^T + N_t\right).
\end{equation}

Let $\Xobs$ be the
obstacle space, so that $\Xfree:= \reals^d\backslash\Xobs$ is the
free space. Let $\Xgoal \subset \Xfree$ and $\xinit \in \Xfree$ be the
goal region and initial state. Given a path cost measure $c$ and letting
$\overline{\x_0,\dots,\x_T}$ denote the continuous curve traced by the
robot's random trajectory (connecting the waypoints $\x_0,\dots,\x_T$)
we wish to solve
\vspace{-0.1truecm}
\begin{quote}{\bf Discretized stochastic motion planning (SMP)}:
\vspace{-0.2truecm}
\begin{equation}
\label{constr}
\begin{split}
\min_{\un(\cdot)} & \quad c(\xn(\cdot)) \\
\text{s.t.} & \quad \prob{\overline{\x_0,\dots,\x_T} \cap \Xobs \neq \varnothing}  \le \alpha\\
& \quad \u_t = \un_t + \du^{\text{LQG}}_t\\
& \quad \x_0 \sim \mathcal{N}(\xn_0, P_0), \quad \x_T \in \Xgoal\\
& \quad \text{Equation  \eqref{eqn:DLQG}}.
\end{split}
\end{equation}
\end{quote}
\vspace{-0.2truecm}
Note that the optimization is still over continuous-time nominal
paths, which we discretize when computing the path collision probability $\prob{\overline{\x_0,\dots,\x_T} \cap \Xobs \neq \varnothing}$.

This formulation is inspired by \cite{FO-CJ-MM:08,JVDB-PA-KG:11} and
represents a compromise between a POMDP formulation involving a minimization over the class of output-feedback control laws, and an open-loop
formulation, in which the state is assumed to evolve in an open loop (i.e., no tracking). This can be justified in two ways. One is that the
general constrained POMDP formulation is vastly more
complex than ours and would require much more computation. The other is that, in
practice, a motion plan is executed in a receding horizon fashion, so that
computing output-feedback policies may
not even be useful, since after a short time-horizon the environment
map may have changed, requiring recomputation anyway. We note that the problem
formulation could be readily generalized to a nonlinear setup and to
any tracking controller (e.g., LQG with extended Kalman filter estimation is essentially
already in the same form)---indeed, one of the key advantages of the MC
approach is that it is able to handle (at least theoretically) such general
versions of the problem. However, in the present paper, we limit our
attention to the aforementioned LQG setup.

In the remainder of the paper, we discuss how to quickly and
consistently (i.e., in a way that is asymptotically exact as the discretization step $\Delta t
\to 0$) estimate the path CP appearing in equation \eqref{constr}, and
then we will employ MCMP to generate approximate solutions to the discretized SMP problem.

\section{The Problem of Computing Path CP}
\label{sec:pathCP}
In general, the key difficulty for planning under uncertainty (provided a probabilistic uncertainty model is given) is to accurately compute path CP. All previous approaches essentially rely on two approaches, namely:
\begin{itemize}
\item {\bf Additive approach}, e.g., \cite{BL-MK-JPH:10}: using Boole's inequality, i.e., %
$
\prob{\cup_i A_i} \leq \sum_i \prob{A_i},
$
by which a path CP is approximated by \emph{summing} pointwise CP$_i$ at a certain number of  waypoints along the path, i.e., $\text{CP} \approx \sum_i \text{CP}_i$.
\item {\bf Multiplicative approach}, e.g.,  \cite{JVDB-PA-KG:11}: a path CP is approximated by \emph{multiplying} the complement of point-wise CP$_i$, specifically $\text{CP} \approx 1 - \prod_i (1- \text{CP}_i)$. %
\end{itemize}

There are three approximations
inherent in both approaches:
\begin{enumerate}[(A)]
\item The path CP is approximated by combining waypoint
  CP$_i$'s. That is, no accounting is made for what happens in
  between waypoints.
    \item The waypoint CP$_i$'s are combined in an approximate
      manner. That is, in general there is a complex high-dimensional
      dependence between collisions at different waypoints, and these
      are not accounted for in either approach. In particular, the
      additive approach treats waypoint collisions as mutually
      exclusive, while the multiplicative approach treats them as
      independent. Since neither mutual exclusivity nor independence
      hold in general, this constitutes another approximation.
\item Each waypoint  CP$_i$ is approximated (e.g., by using a
  nearest obstacle). This is usually done because integrating a
  multivariate density over an intersection of half-planes (defining
  the obstacle set) can be quite computationally expensive.
\end{enumerate}

\begin{figure}[h]
\centering
		\subfigure[]{\label{fig:CP_1}	\includegraphics[width=0.3\textwidth]{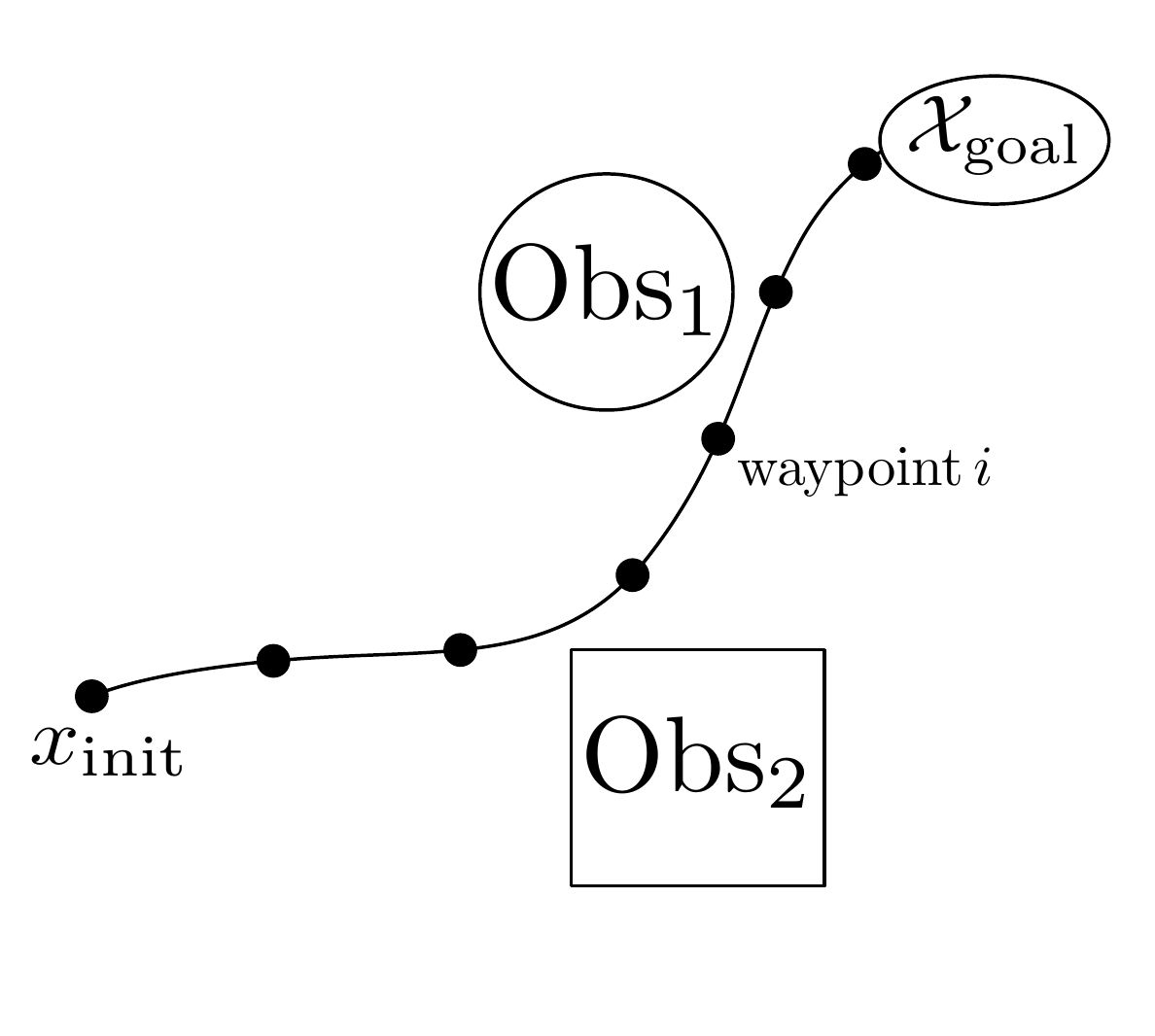}} \qquad
		\subfigure[]{\label{fig:CP_2}
\includegraphics[width=.3\textwidth]{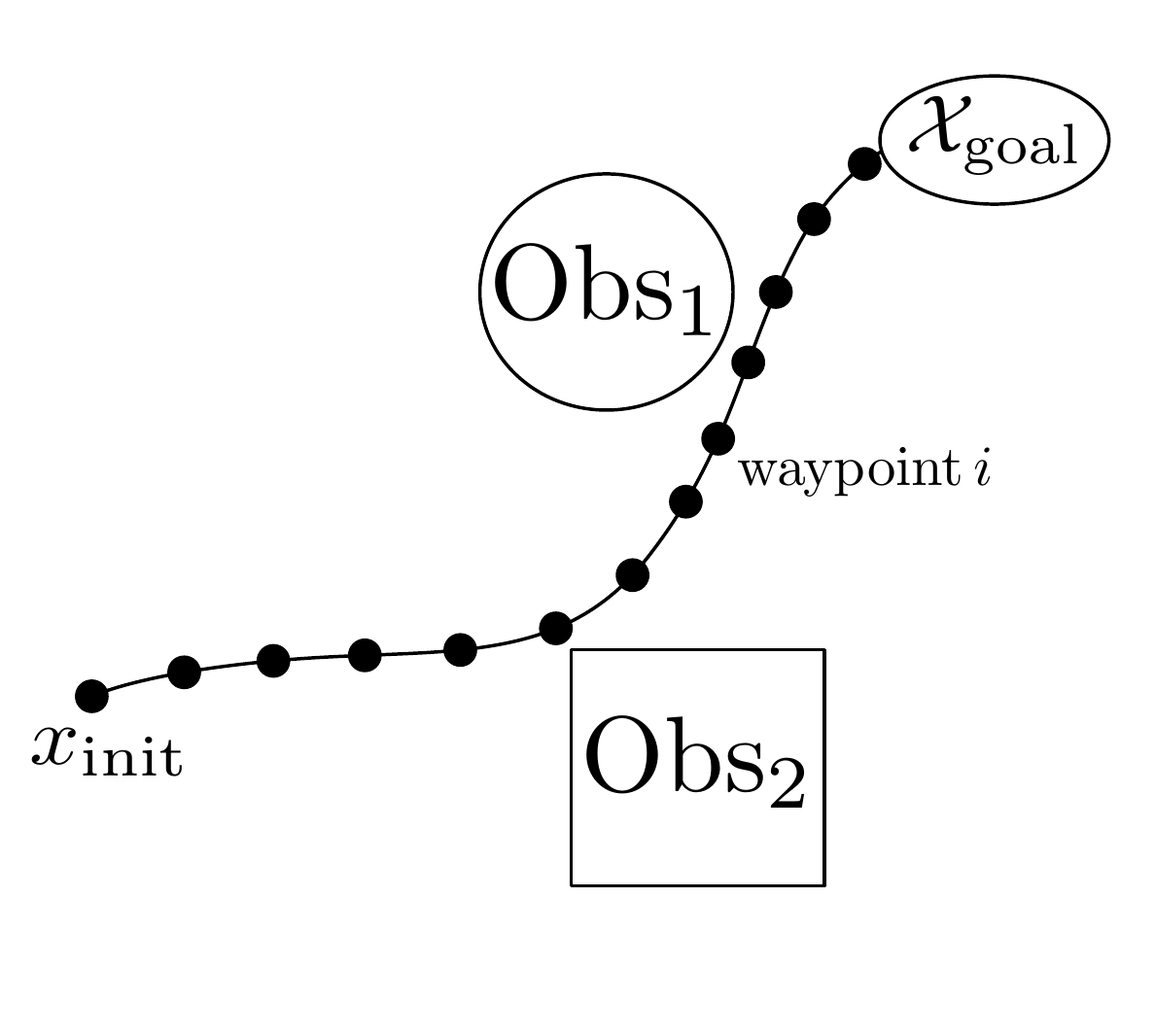}}
\vspace{-0.3truecm}
\caption{Illustration of the interplay between approximations (A) and
  (B). In (a), there are not enough waypoints to properly characterize
  the pathwise CP, while in (b), the waypoints may be too close to not
  account for their dependence.}
\label{fig:CP_resolution}
\vspace{-0.7cm}
\end{figure}

\begin{figure}[t]
\centering
    \subfigure[]{\label{fig:CP_path} \includegraphics[width=.36\textwidth]{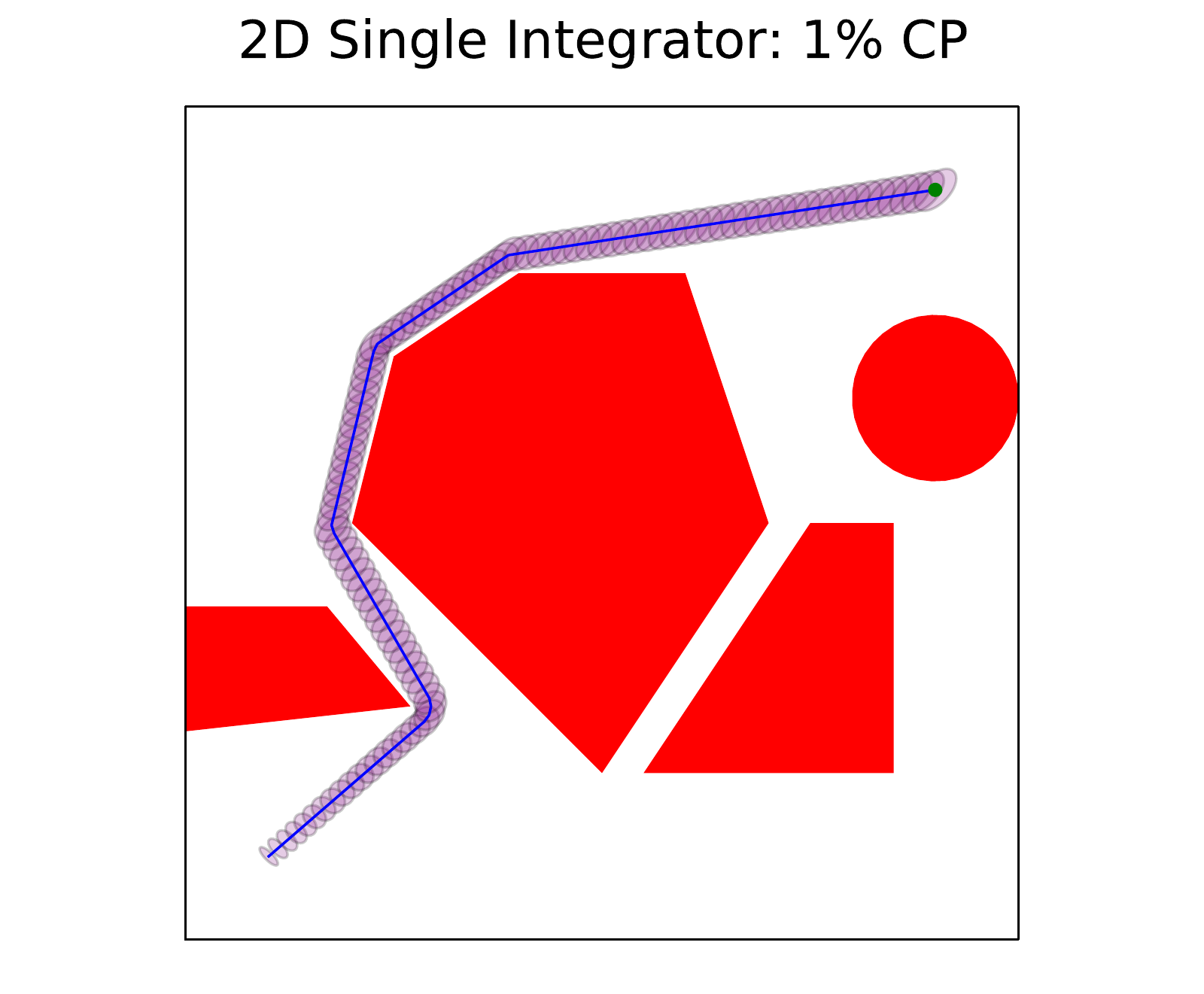}} \qquad
    \subfigure[]{\label{fig:CP_plot1} \includegraphics[width=.36\textwidth]{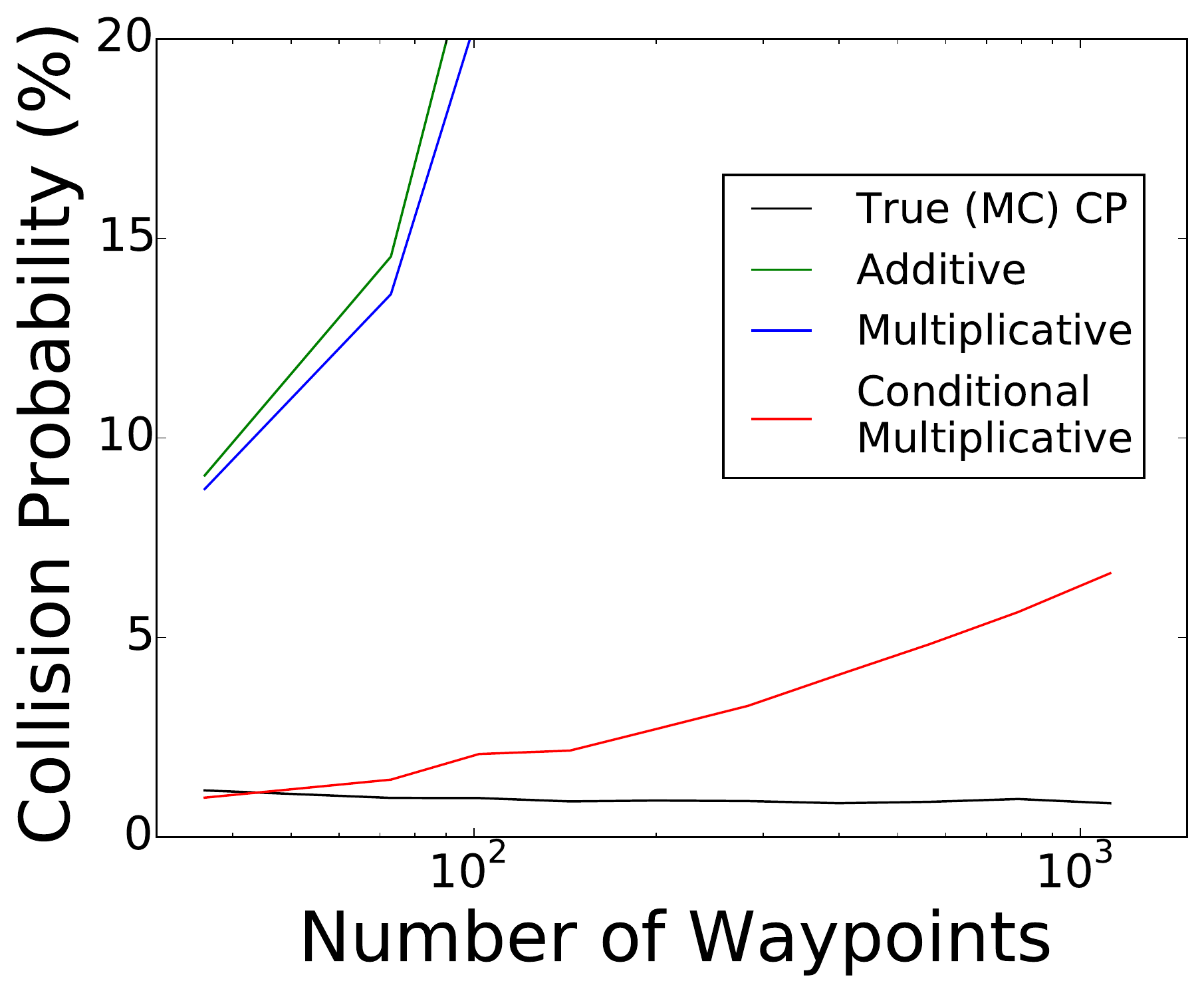}}
		\subfigure[]{\label{fig:CP_path} \includegraphics[width=.36\textwidth]{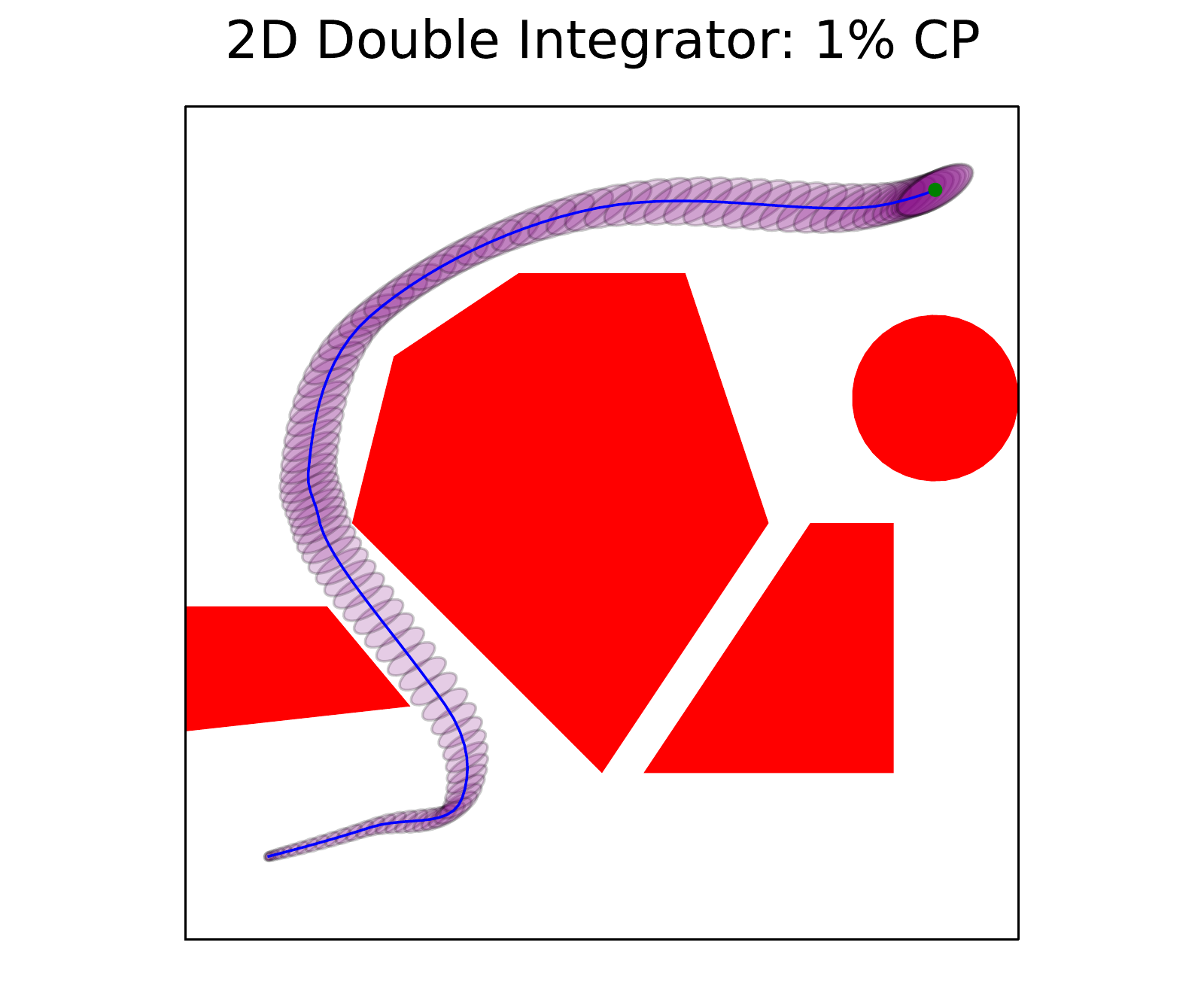}} \qquad
		\subfigure[]{\label{fig:CP_plot2} \includegraphics[width=.36\textwidth]{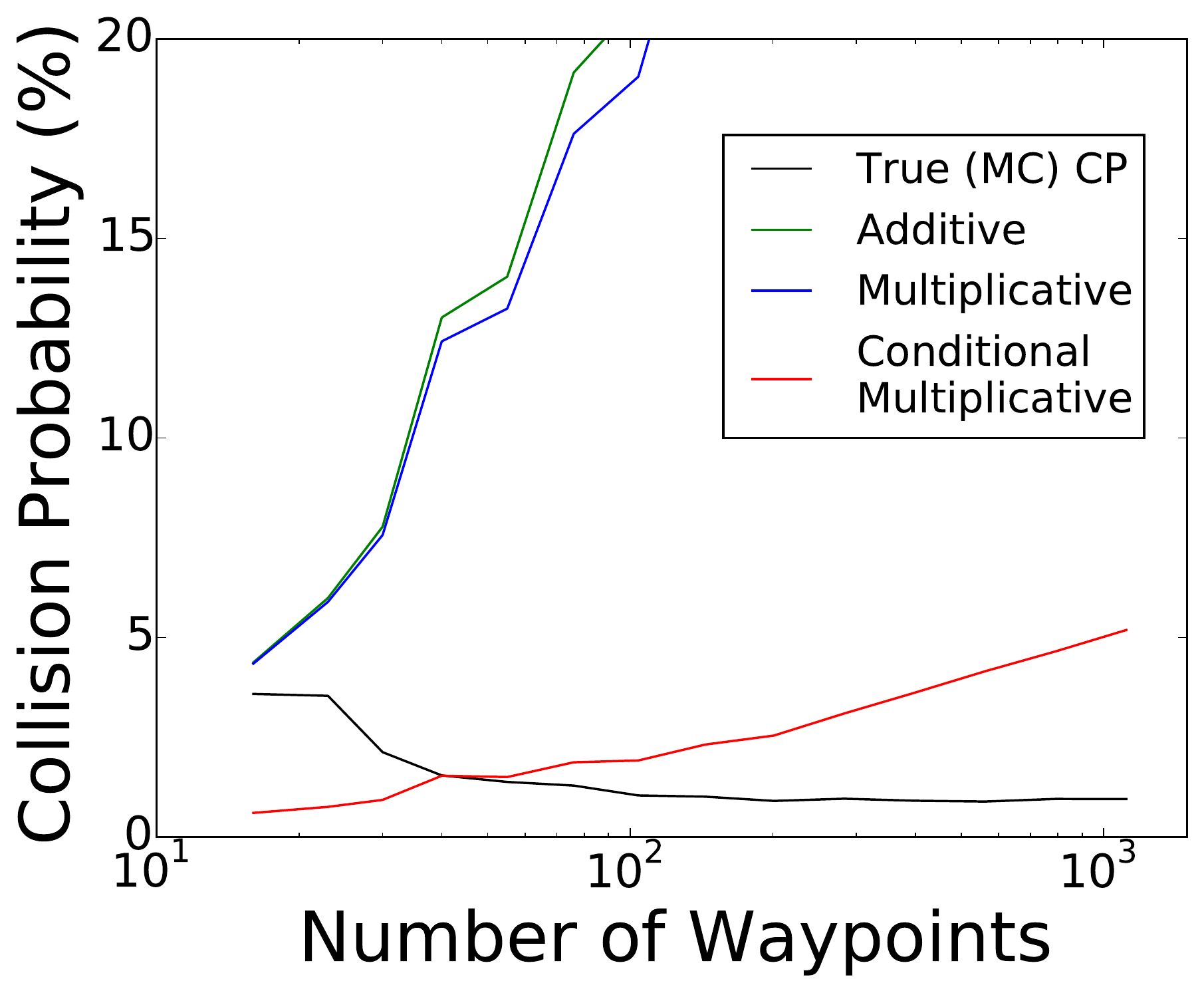}}
\vspace{-0.3truecm}
\caption{Illustration of path CP approximation schemes for two
  robotic systems where the true path CP is around
  1\% under the continuous controller. In (a) and (c), the blue curve
  represents the nominal path, the red boxes are the obstacles, and
  the purple ellipses represent 95\% pointwise marginal confidence
  intervals at individual waypoints at discretizations with 102 and 104
  points respectively. Panels (b) and (d) show the collision probability estimated by each approximation scheme as a function
  of the number of waypoints.
  Approximation
  (C) in all approaches is matched to their respective papers
  (additive: \cite{BL-MK-JPH:10, GSA-BDL-JMJ-NR-JPH:13, MK-IP:13,
    BDL-SK-JPH:13}, multiplicative: \cite{JVDB-PA-KG:11}, conditional
  multiplicative: \cite{SP-JVDB-RA:12, SW-LGT-JVDB-RA:13, WL-MA:14}).}
\label{fig:CP_results}
\vspace{-0.3truecm}
\end{figure}

A fundamental limitation in both approaches comes from the interplay between approximations (A) and (B). Specifically, while approximation (A) improves with higher-resolution waypoint placement
along the path, approximation (B) actually gets
worse, see Figure~\ref{fig:CP_resolution}. In
Figure~\ref{fig:CP_1}, although Obs$_2$ comes very close to the path,
it does not come very close to any of the waypoints, and thus the
pathwise CP will not be properly accounted for
by just combining pointwise CPs. In Figure~\ref{fig:CP_2}, the
waypoints closest to Obs$_1$ will have highly-correlated CPs, which
again is not accounted for in either the additive or multiplicative
approaches. For the linear Gaussian setting considered here, as the
number of waypoints along a fixed path goes to infinity, the path CP
estimate from the additive approach actually tends to $\infty$, while
that of the multiplicative approach tends to 1, \emph{regardless} of
the true path CP. To see this, note that for any fixed path, there
exists a positive number $\epsilon>0$ such that CP$_i$  is larger than
or equal to $\epsilon$ for \emph{any} point on the path. Therefore,
\begin{equation}
\label{prob}
\sum_{i=1}^k \text{CP}_i  \ge k\, \epsilon 
\stackrel{k\rightarrow\infty}{\longrightarrow} \infty, \qquad \; 1 \ge 1-\prod_{i=1}^k (1-\text{CP}_i)  \ge 1-(1- \epsilon )^k
\stackrel{k\rightarrow\infty}{\longrightarrow} 1,
\end{equation}
where $k$ is the number of waypoints. In other words, both approaches
are asymptotically tautological, as they upper-bound a probability
with a number greater than or equal to one. An important consequence of this is
that as the number of waypoints approaches infinity, either approach
would deem \emph{all} possible paths infeasible with respect to
\emph{any} fixed non-trivial path CP constraint. This point is emphasized in Figure
\ref{fig:CP_results}, which compares true path CP to approximations
computed using the additive and multiplicative approaches
for two different paths, as a function of the number of waypoints
along the path. Off the plotted area, the additive approach passes through an approximate probability of
1 and continues to infinity,
while the multiplicative approach levels off at 1. Even with few waypoints, both approaches are off by
hundreds of percent. The overly
conservative nature of the multiplicative approach has been recognized
in \cite{SP-JVDB-RA:12}, where the authors replace approximate
marginal pointwise CPs in the multiplicative approach with approximate
pointwise CPs \emph{conditional} on the previous waypoint being
collision-free. While this is a first-order improvement on the
standard approaches, the conditional pointwise probabilities are quite complex
but are approximated by Gaussians for computational reasons, with the
result that their approximate path CPs can still be off by many
multiples of the true value, especially for small path CPs (which
will usually be the relevant ones). The red curve in
Figure~\ref{fig:CP_plot1} and \ref{fig:CP_plot2} shows that the
approximation of \cite{SP-JVDB-RA:12}, while a substantial
improvement over the alternatives, can still be off by factors of 5
or more, and the discrepancy appears to be increasing steadily with
the number of waypoints.

A few comments are in order.
First, there is nothing pathological in the example in
Figure~\ref{fig:CP_results}, as similar results are obtained with
other obstacle configurations and in higher dimensions.
Second, we note that approximations such
as these may be very useful for \emph{unconstrained} problems that
penalize or minimize path CP, since they may measure the
\emph{relative} CP between paths well, even if they do not agree with
the true path CP in absolute terms. However, to address the
chance-constrained SMP, one needs an accurate (in absolute terms) and
fast method to estimate path CP, which is one of the key contributions
of this paper.
Third, the additive approach is guaranteed to
be conservative with respect to approximation (B). That is, ignoring 
(A) and (C) (the latter of which can also be made
conservative), the additive approach will produce an overestimate of
the path CP. Although this can result in high-cost paths or even
problem infeasibility, it is at least on the safe side. This guarantee
comes at a cost of extreme conservativeness, and the two
less-conservative multiplicative approaches have no such guarantee for
any finite number of waypoints.
Fourth, the
limits in equation~\eqref{prob} apply to any uncertainty model
supported on the entire configuration space, and even to
bounded uncertainty models so long as the path in question has a
positive-length segment of positive pointwise CP.

In the next section, we will present a MC approach that addresses all
three approximations (A)-(C) stated earlier. Specifically, for (A), although
collisions can truly be checked along a continuous path only for
special cases of obstacles, Monte Carlo simply checks for collisions
along a sampled path and thus can do so at arbitrary resolution, regardless of the
resolution of the actual waypoints, so approximation (A) for MC has no
dependence on waypoint resolution.  For (B), the high-dimensional
joint distribution of collisions at waypoints along the path is
automatically accounted for when sampling entire realizations of the
tracking controller. And for (C), since MC only has to check for
collisions at specific points in space, no multivariate density
integration needs to be done.

\section{Variance-Reduced Monte Carlo for Computing Pathwise CP}
\label{sec:main}
\subsection{Control Variates}
\label{sec:main.cv}
As discussed in Section~\ref{sec:cv.background}, a good control variate $h$ for $f$
should have a known (or rapidly computable) expected value, and should
be highly correlated with $f$. As mentioned in the previous section,
existing probability-approximation methods, while not accurate in
absolute terms,
can act as very good proxies for CP in that they resemble a monotone
function of the CP. Coupled with the fact that such approximations are
extremely fast to compute, they make ideal candidates
for $h$.

Since even individual waypoint CPs are expensive to compute exactly for all but the simplest obstacle sets, we approximate the obstacle set locally as a union of half-planes, similar to \cite{SP-JVDB-RA:12}. For each waypoint $\x^{\text{nom}}_t$
along the nominal path, we compute the closest obstacle points $\z_t^{(i)}$ and their corresponding obstacle half-planes such that none of these points are occluded by each others' half-planes. ``Close'' is measured in terms of the Mahalanobis distance
defined by the covariance matrix of the robot state at that waypoint, and the
obstacle half-planes are defined as tangent to the multivariate Gaussian density contour
at each close point. Mathematically this corresponds to at most one point per convex obstacle region $\Xobs^{(i)}$ (with $\Xobs = \bigcup_{i=1}^M \Xobs^{(i)}$), i.e.,
\[
\z_t^{(i)} = \argmin_{\mathbf{a} \in \Xobs^{(i)}} (\mathbf{a} - \x^{\text{nom}}_t)^T \Sigma_t^{-1} (\mathbf{a} - \x^{\text{nom}}_t).
\]
We then approximate the pointwise probability of collision by the probability of crossing any one of these half-planes; this probability is approximated in turn by Boole's bound so that an expectation is simple to compute. That is, we define $h_{ti}(\bs{X})$ to be the indicator that $\x^{\text{nom}}_t$ crosses the $\z_t^{(i)}$ obstacle half-plane, and define $h(\bs{X}) = \sum_{t,i} h_{ti}(\bs{X})$.
We note that considering multiple close obstacle points, as opposed to only the closest one, is important when planning in tight spaces with obstacles on all sides. Correlations between $h$ and $f$ in testing were regularly around 0.8.

\subsection{Importance Sampling}
\label{sec:main.is}
From a statistical standpoint, the goal in selecting the importance
distribution $Q$ is to make the pathwise CP sampled under $Q$ on the order of 1, while
keeping the colliding paths sampled from $Q$ as likely as possible 
under the nominal distribution $P$. From a computational standpoint,
we want $Q$ to be fast to sample from and for the likelihood ratio $P/Q$ to be
easy to compute. Our method for importance sampling constructs $Q$ as a mixture
of sampling distributions $Q_0,...,Q_K$---one for each close obstacle point $\z_t^{(i)}$ along the nominal
trajectory. The intent of distribution $Q_{ti}$ is to sample a path that is likely to collide
with the obstacle set at waypoint $t$. We accomplish this by shifting the means of the
noise distributions $\n_s$, $0 \leq s \leq t$, leading up to time $t$ so that
$\mathbb{E}_{Q_{ti}}[\dx_t] = \z_t^{(i)} - \x_t^\text{nom}$. To
minimize
likelihood ratio $P(\bs{X}) / Q_{ti}(\bs{X})$ variance, we distribute the shift in the most likely
manner according to Mahalanobis distance.
This amounts,
through Equation \eqref{eqn:uncertaintyevol},
to solving the least-squares problem
\begin{equation}\label{eqn:ISopt}
\begin{aligned}
\min_{\dm_0,\dots,\dm_t} & \quad \sum_{s=0}^t \dm_s^T N_s^{-1} \dm_s \\
s.t. & \quad \sum_{s=0}^t \begin{bmatrix} I & 0 \end{bmatrix}
\left(\prod_{r=0}^{t-s-1} M_{t-r} \right)\dm_s = \z_t^{(i)} - \x_t^\text{nom}
\end{aligned}
\end{equation}
and sampling the noise as $\tilde \n_s \sim \mathcal{N}\left(\dm_s, N_s\right)$
for $0 \leq s \leq t$.

We weight the full mixture IS distribution, with $\theta = \mathbb{E}[h(\bs{X})]$, as
\[
Q = \sum_{t, i} \left(\frac{\mathbb{E}[h_{ti}(\bs{X})]}{\theta}\right) Q_{ti}.
\]
That is, the more likely it is for the
true path distribution to collide at $t$, the more likely
we are to sample a path pushed toward collision at $t$.

\subsection{Combining the Two Variance-Reduction Techniques}
Due to space limitations, we do not discuss the full details of combining CV and IS here, but simply state the final combined procedure in Algorithm~\ref{mc}.
\vspace{-0.4truecm}
\begin{algorithm}[h]
\caption{Monte Carlo Path CP Estimation}
\label{mc}
\begin{algorithmic}[1]
\REQUIRE Nominal distribution $P$, control variate $h$ as in Section~\ref{sec:main.cv}, $\theta:=
\mathbb{E}_P[h(\bs{X})]$, importance distribution
$Q$ as in Section~\ref{sec:main.is}, number of samples $m$
\STATE Sample $\{\tilde{\bs{X}}^{(i)}\}_{i=1}^m$ i.i.d. from $Q$
\STATE Denoting the likelihood ratio $L(\tilde{\bs{X}}^{(i)}) := P(\tilde{\bs{X}}^{(i)}) / Q(\tilde{\bs{X}}^{(i)})$, compute
\begin{align*}
\hat{p}_Q &= \frac{1}{m}\sum_{i=1}^m f(\tilde{\bs{X}}^{(i)})
  L(\tilde{\bs{X}}^{(i)}), \qquad \quad
\hat{\theta}_Q = \frac{1}{m}\sum_{i=1}^m h(\tilde{\bs{X}}^{(i)})
  L(\tilde{\bs{X}}^{(i)}),\\
\hat{\beta}_Q &= \frac{\sum_{i=1}^m \left(
f(\tilde{\bs{X}}^{(i)})L(\tilde{\bs{X}}^{(i)}) -
\hat{p}_Q\right) \left(h(\tilde{\bs{X}}^{(i)})L(\tilde{\bs{X}}^{(i)})
- \hat{\theta}_Q\right)}{\sum_{i=1}^m
\left(h(\tilde{\bs{X}}^{(i)})L(\tilde{\bs{X}}^{(i)}) - \hat{\theta}_Q\right)^2},\\
\hat{p}_{Q, \hat{\beta}_Q} &= \frac{1}{m}\sum_{i=1}^m
f(\tilde{\bs{X}}^{(i)})L(\tilde{\bs{X}}^{(i)}) - \hat{\beta}_Q
  h(\tilde{\bs{X}}^{(i)})L(\tilde{\bs{X}}^{(i)}) + \hat{\beta}_q\theta\\
\hat{V}_{Q,\hat{\beta}_Q} &= \frac{1}{m^2}\sum_{i=1}^m
\left( f(\tilde{\bs{X}}^{(i)})L(\tilde{\bs{X}}^{(i)}) -
\hat{p}_{Q,\hat{\beta}_Q}- \hat{\beta}\left(h(\tilde{\bs{X}}^{(i)})L(\tilde{\bs{X}}^{(i)})
- \theta\right)\right)^2
\end{align*}
\RETURN $\hat{p}_{Q, \hat{\beta}_Q}$, $\hat{V}_{Q,\hat{\beta}_Q}$
\end{algorithmic}
\end{algorithm}
\vspace{-0.4truecm}

\section{MCMP Algorithm}
\label{sec:algo}
With an algorithm for path CP estimation in hand, we now incorporate
it into a simple scheme for generating high-quality paths subject to a
path CP constraint. Algorithm~\ref{mcmp} describes the Monte Carlo
Motion Planning (MCMP) algorithm in pseudocode.
\begin{algorithm}[h]
\caption{Monte Carlo Motion Planning}
\label{mcmp}
\begin{algorithmic}[1]
\REQUIRE Maximum inflation $I_{\text{max}}$ (e.g. configuration space
diameter), %
minimum inflation $I_{\text{min}}$ (e.g. 0), number of
bisection steps $r$, path CP constraint $\alpha$
\FOR{$i = 1:r$}
\STATE Compute an (approximately) optimal path $\hat{\sigma}$ using,
e.g., an asymptotically optimal sampling-based motion planning (SBMP) algorithm, for the deterministic
version of the problem with the obstacles inflated by
${(I_{\text{min}}+I_{\text{max}})/2}$ \label{line:plan}
\STATE Compute a MC estimate $\hat{p}$ of the CP of $\hat{\sigma}$
(set $\hat{p}=0$ if the previous step fails to find a feasible solution)\label{line:mc}
\IF{$\hat{p}>\alpha$}
\STATE $I_{\text{min}} = (I_{\text{min}}+I_{\text{max}})/2$
\ELSE
\STATE $I_{\text{max}} = (I_{\text{min}}+I_{\text{max}})/2$
\ENDIF
\ENDFOR
\RETURN $\hat{\sigma}$
\end{algorithmic}
\end{algorithm}

The idea of MCMP is simple: solve the deterministic motion
planning problem with inflated obstacles to make the resulting path
safer, and then adjust the inflation so that the path is exactly as
safe as desired. Note that in line~\ref{line:mc} of
Algorithm~\ref{mcmp}, MC could be replaced by any of the
approximations from Section~\ref{sec:pathCP}, but the output would
suffer in quality. In the case of the multiplicative approaches, the
CP may be underestimated, in which case the safety constraint will be
violated. More commonly (for a reasonable number of waypoints), for
both additive and multiplicative approaches, the CP may be
substantially overestimated. Although the resulting path will not
violate the safety constraint, it will be inefficient in that it will
take a costlier path than needed (or than the one returned by using MC
CP estimation) in order to give the obstacles a wider berth than
necessary. Another possibility is that the obstacle inflation needed
to satisfy the conservative safety constraint actually closes off all
paths to the goal, rendering the problem infeasible, even
if it may have been feasible using MC estimation. 

It is worth pointing out that the tails, or probability of extreme
values, of the Gaussian distribution fall off very rapidly, at a
double-exponential rate. For instance, the $0.01^{th}$ percentile of a
Gaussian distribution is only about 20\% farther from the mean than
the $0.1^{th}$ percentile. In the Gaussian framework of this paper,
this means that a path that already has a small CP can make its CP
much smaller by only shifting slightly farther from the
obstacles. Thus although the additive or multiplicative approximations
may overestimate the pathwise CP by hundreds of percent, the cost
difference between using them in line~\ref{line:mc} of
Algorithm~\ref{mcmp} and using MC in line~\ref{line:mc} of
Algorithm~\ref{mcmp} may not be nearly so
drastic. However, it can be if the increased obstacle inflation
required closes off an entire homotopy class, or renders the problem
infeasible altogether.

\section{Numerical Experiments}
\label{sec:simulations}

We implemented variance-reduced path CP estimation and MCMP in Julia \cite{JB-SK-VBS-AE:12} for numerical experiments on a range of linear dynamical systems and obstacle sets, run using a Unix operating system with a
2.0 GHz processor and 8 GB of RAM.
Many implementation details and tuning parameters have been omitted in the discussion below;
the code for these results may be accessed at \url{https://github.com/schmrlng/MCMP-ISRR15}.

\begin{figure}[t]
\centering
    \subfigure[]{\label{fig:mcmp_1} \includegraphics[width=0.32\textwidth,
                  trim=2.5cm 0cm 1.2cm 0cm, clip=true]{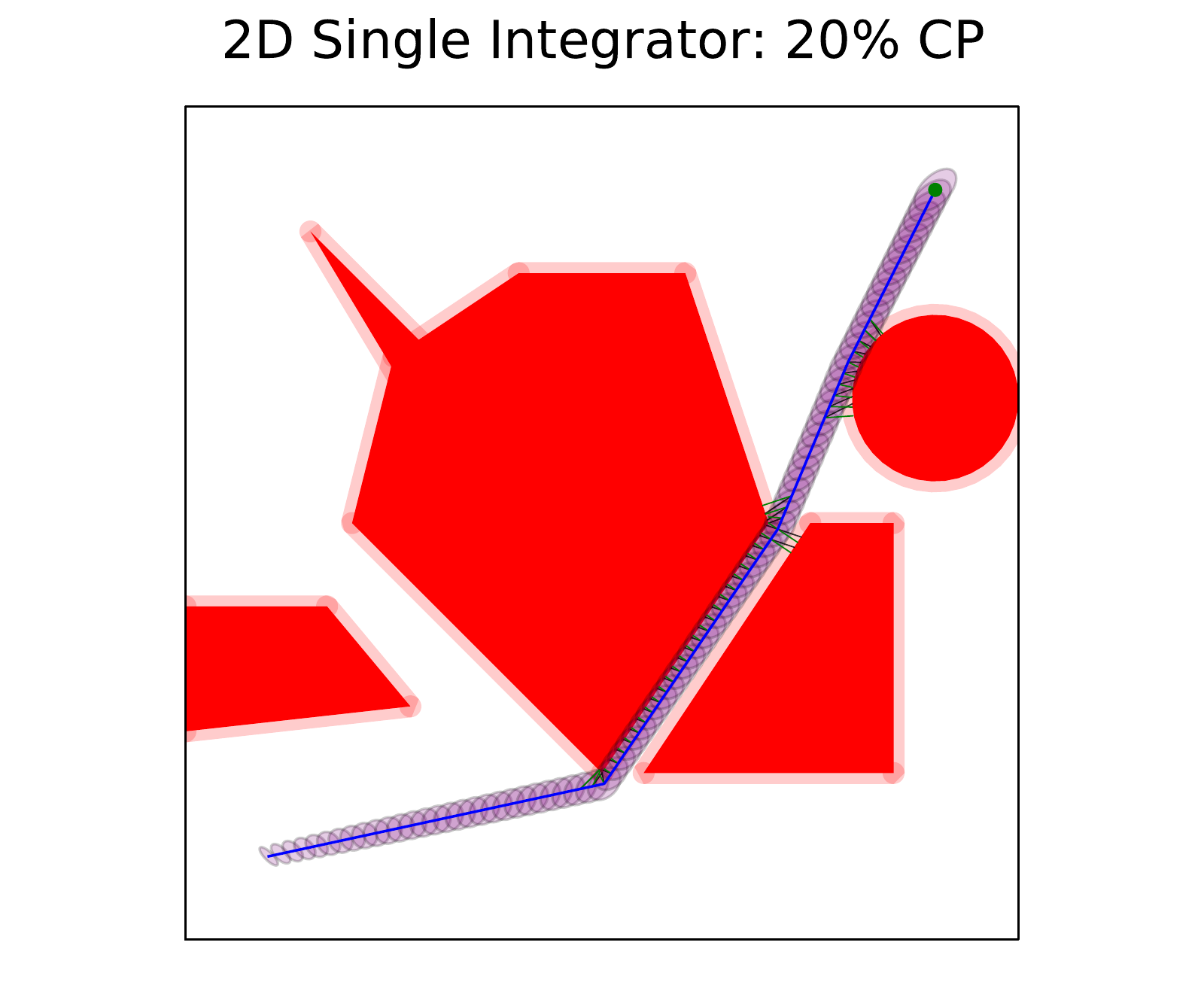}}
    \subfigure[]{\label{fig:mcmp_2}
\includegraphics[width=.32\textwidth, trim=2.5cm 0cm 1.2cm 0cm, clip=true]{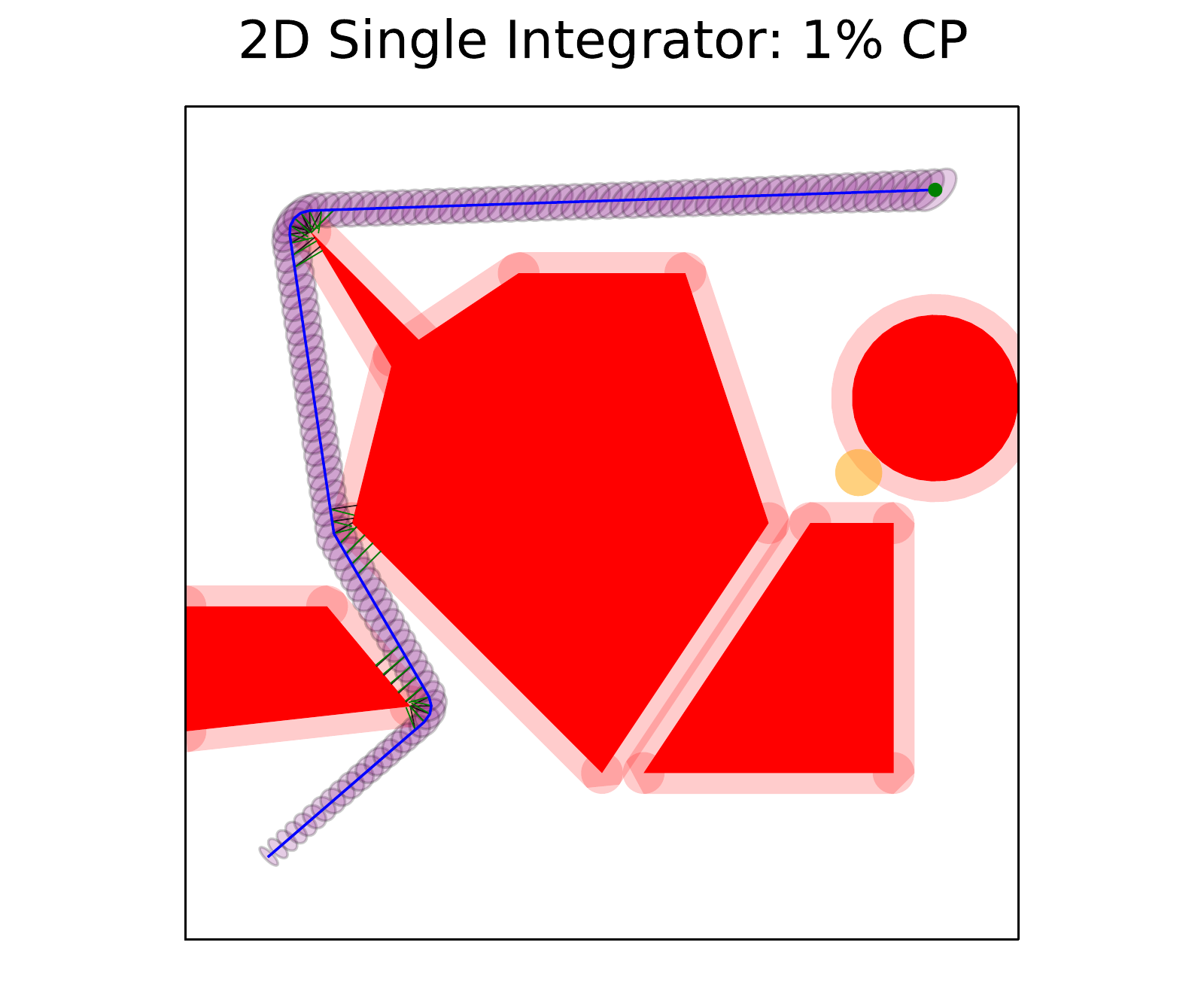}}
    \subfigure[]{\label{fig:mcmp_3}
\includegraphics[width=.32\textwidth, trim=2.5cm 0cm 1.2cm 0cm, clip=true]{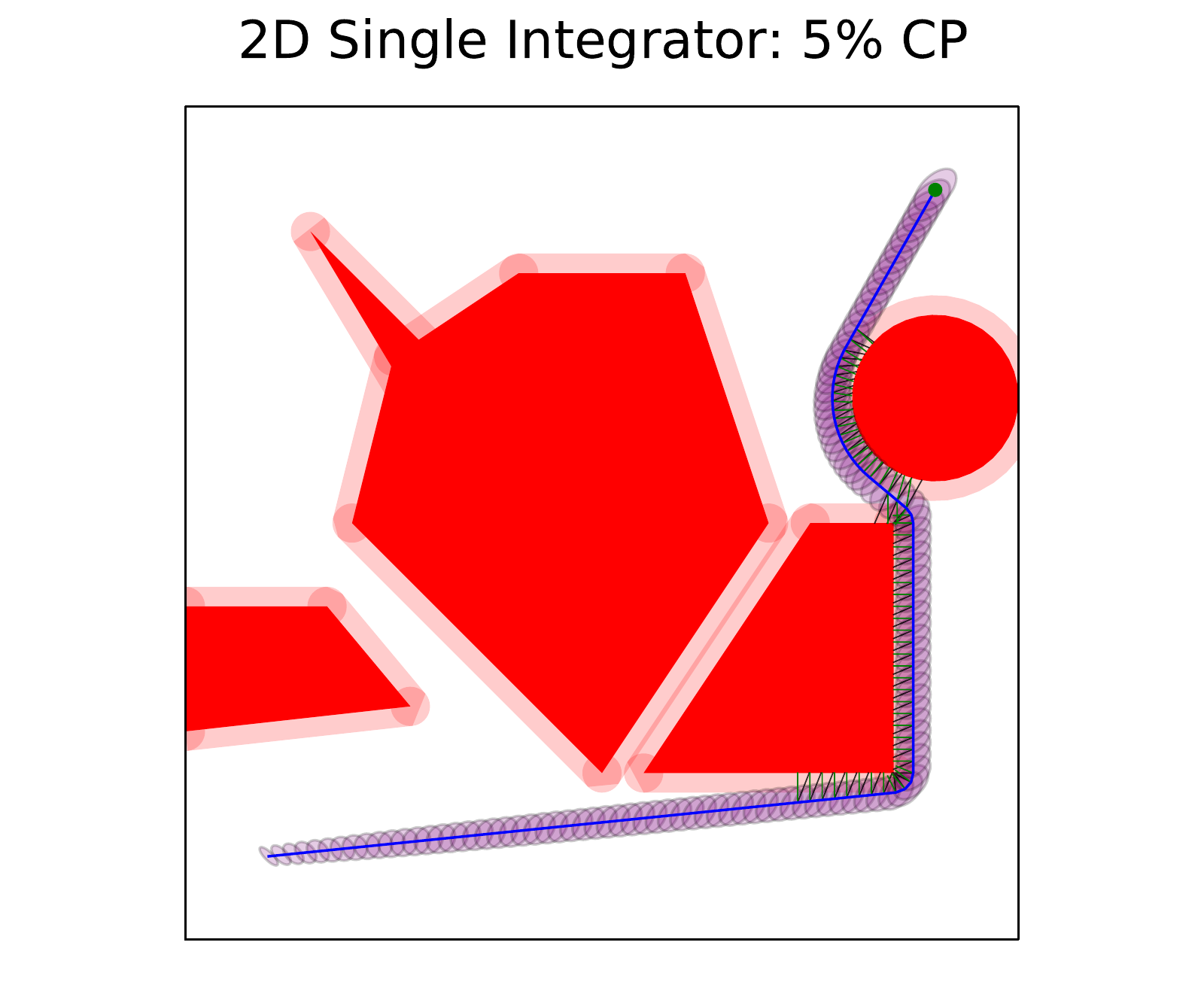}}
\vspace{-0.3truecm}
\caption{Illustration of the MCMP algorithm output given a range of target path CPs
  for a 2D single integrator system. For these uncertainty parameters, we see that
  the precise safety tolerance value (between 1\%--20\%) will correspond to a nominal
  solution in one of three distinct homotopy classes. The orange obstacle in (b)
  is added by the ``block and backtrack'' modification, discussed in Section~\ref{sec:simulations}, to the basic MCMP bisection Algorithm~\ref{mcmp}. The black and green lines denote the close obstacle points and vectors defining their half planes respectively; only the pruned set is depicted.
  }
\label{fig:mcmp}
\vspace{-0.2truecm}
\end{figure}

Figures~\ref{fig:CP_results} and \ref{fig:mcmp} display some example results for single integrator ($\dot x = u$) and double integrator ($\ddot x = u$) systems in a two-dimensional workspace, and Table~\ref{tab:SSresults} summarizes a range of statistics on algorithm performance in three-dimensional workspaces as well. The deterministic planning step (Algorithm~\ref{mcmp}, line~\ref{line:plan}) was accomplished using the differential \FMT algorithm \cite{ES-LJ-MP:15b} on a fixed set of nodes. By caching nearest-neighbor and distance data for these nodes (Offline Planning), the total replanning time over all inflation factors (Online Planning, essentially consisting only of collision checking) was significantly reduced. For the single integrator systems 2D SI and 3D SI, the planning problem is equivalent to geometric planning, which allowed us to apply the ADAPTIVE-SHORTCUT rubber-band-style heuristic for smoothing planned paths \cite{DH:00}. Applying this smoothing heuristic ensures that the path CP varies continuously with inflation factor within a path homotopy class. Between homotopy classes the CP may be discontinuous as a function of inflation factor. If increasing the inflation factor increases the CP discontinuously, the bisection process is not affected; otherwise if the CP decreases (e.g. Figure~\ref{fig:mcmp} (b) and (c)---the CPs are 0.3\% and 1.5\% respectively around the inflation factor which closes off the (c) route) the MCMP bisection algorithm may get stuck before reaching the target CP $\alpha$ (e.g. 1\% in the case of Table~\ref{tab:SSresults} row 2). To remedy this issue, we implemented a ``block and backtrack'' modification to Algorithm~\ref{mcmp} which blocks off the riskier homotopy class with an obstacle placed at its waypoint most likely to be in collision, and then resets the bisection lower bound for the inflation factor. This results in increased computation time, but returns a path with the goal CP in the end.

We did not implement any smoothing procedure for the double integrator systems. Each nominal trajectory is selected as a concatenation of local steering connections, subject to variance in the placement of the finite set of planning nodes. In practice, this means that path CP is piecewise constant, with many small discontinuities, as a function of inflation factor. If the bisection procedure terminates at an interval around the desired CP, we choose the path satisfying the safety constraint: this explains the mean True CP below the goal value in Table~\ref{tab:SSresults} rows 4 and 5.

In order to speed up the Monte Carlo CP estimation, we prune the set of close obstacle points $\z_t^{(i)}$ so that the ones that remain are expected to have their term in the mixture distribution $Q$ sampled at least once. We note that this style of pruning does not bias the results; it only affects computation time and estimator variance. Additionally, since during each MCMP run we only use the CP estimate for bisection, we also save time by terminating the estimation procedure early when estimated estimator variance suggests we may do so with confidence.

\begin{table*}[t]
\centering
\begin{tabular}{|l|r|r|r|r|r|r|r|}
\hline
 & \multicolumn{1}{|c}{Goal }
 & \multicolumn{1}{|c}{Offline}
 & \multicolumn{1}{|c}{Online}
 & \multicolumn{1}{|c}{MC}
 & \multicolumn{1}{|c}{Discretization}
 & \multicolumn{1}{|c|}{Bisection}
 & \multicolumn{1}{|c|}{MC}\\
 & \multicolumn{1}{|c}{CP (\%)}
 & \multicolumn{1}{|c}{Planning (s)}
 & \multicolumn{1}{|c}{Planning (s)}
 & \multicolumn{1}{|c}{Time (s)}
 & \multicolumn{1}{|c}{Points}
 & \multicolumn{1}{|c|}{Iterations}
 & \multicolumn{1}{|c|}{Particles}\\ \hline
2D SI (A) & 1 &    0.25 $\pm$ 0.03 & 1.25 $\pm$ 0.24 & 2.64 $\pm$ 0.83 & 102.5 $\pm$ 0.8 & 6.3 $\pm$ 1.5 & 2085 $\pm$ 686\\ \hline
2D SI (B) & 1 &    0.27 $\pm$ 0.04 & 2.48 $\pm$ 0.77 & 4.65 $\pm$ 1.70 & 116.5 $\pm$ 0.7 & 13.3 $\pm$ 4.4 & 2955 $\pm$ 1052\\ \hline
3D SI & 1 &        0.35 $\pm$ 0.03 & 1.95 $\pm$ 0.76 & 3.00 $\pm$ 0.89 & 83.6 $\pm$ 1.4 & 6.3 $\pm$ 3.1 & 1667 $\pm$ 764\\ \hline
2D DI & 1 &        6.64 $\pm$ 0.14 & 2.86 $\pm$ 0.98 & 5.82 $\pm$ 2.33 & 107.7 $\pm$ 6.0 & 8.6 $\pm$ 2.9 & 2383 $\pm$ 952\\ \hline
3D DI & 1 &        20.90 $\pm$ 1.11 & 6.27 $\pm$ 2.40 & 7.45 $\pm$ 3.75 & 71.7 $\pm$ 10.8 & 7.8 $\pm$ 3.3 & 2117 $\pm$ 938\\ \hline
\end{tabular}

\vspace{.25cm}

\begin{tabular}{|l|r|r|r|r|r|r|}
\hline
 & \multicolumn{1}{|c}{Goal}
 & \multicolumn{1}{|c}{Nominal}
 & \multicolumn{1}{|c}{True (MC)}
 & \multicolumn{1}{|c}{Additive}
 & \multicolumn{1}{|c}{Multiplicative}
 & \multicolumn{1}{|c|}{Cond. Mult.}\\
 & \multicolumn{1}{|c}{CP (\%)}
 & \multicolumn{1}{|c}{Path Cost}
 & \multicolumn{1}{|c}{CP (\%)}
 & \multicolumn{1}{|c}{Estimate (\%)}
 & \multicolumn{1}{|c}{Estimate (\%)}
 & \multicolumn{1}{|c|}{Estimate (\%)}\\ \hline
2D SI (A) & 1 &  1.47 $\pm$ 0.00 & 1.01 $\pm$ 0.06 & 22.67 $\pm$ 2.39 & 20.35 $\pm$ 1.92 & 2.04 $\pm$ 0.20\\ \hline
2D SI (B) & 1 &  1.69 $\pm$ 0.01 & 1.00 $\pm$ 0.06 & 12.88 $\pm$ 3.72 & 12.10 $\pm$ 2.97 & 1.37 $\pm$ 0.26\\ \hline
3D SI & 1 &      1.28 $\pm$ 0.03 & 1.00 $\pm$ 0.06 & 47.48 $\pm$ 7.98 & 38.84 $\pm$ 5.51 & 2.15 $\pm$ 0.23\\ \hline
2D DI & 1 &      7.20 $\pm$ 0.43 & 0.67 $\pm$ 0.27 & 15.04 $\pm$ 8.97 & 13.78 $\pm$ 7.59 & 1.39 $\pm$ 0.68\\ \hline
3D DI & 1 &      9.97 $\pm$ 1.61 & 0.66 $\pm$ 0.33 & 12.26 $\pm$ 5.88 & 11.68 $\pm$ 5.43 & 0.59 $\pm$ 0.32\\ \hline
\end{tabular}
\caption{{\bf (MCMP in various state spaces).} Results averaged over 400 MCMP runs: 20 runs each for 20 SBMP sample sets. SI and DI denote single and double integrator respectively. We aim to minimize arc-length for the SI systems, and a mixed time/control energy cost for the DI systems. 2D SI (A) refers to the obstacle set in Figure~\ref{fig:CP_results}, and 2D SI (B) refers to the obstacle set in Figure~\ref{fig:mcmp}.}
\label{tab:SSresults}
\vspace{-0.5truecm}
\end{table*}

From Table~\ref{tab:SSresults} we see that MCMP run times approach real time in a range of state spaces from 2--6 dimensions, on the order of 5--10 seconds total, excluding planning computation that may be cached offline. This is accomplished even at a level of tracking discretization sufficient to approximate continuous LQG. Planning time and probability estimation time are similar in magnitude, indicating that the MC portion of MCMP is not significantly holding back algorithm run time compared to a faster approximation scheme, even in this single processor implementation. Computing the Monte Carlo path simulations (MC Particles) in parallel could greatly reduce that time. We note that the few thousand simulations required in total by MCMP would not be enough to certify, using simple Monte Carlo, that a path CP is within the interval $(0.9\%, 1.1\%)$ even once, which highlights the effectiveness of our proposed estimator variance reduction techniques.

As can be seen from the simulations, the accuracies of the additive, multiplicative, and conditional multiplicative approximations vary over problems and parameters, even occasionally being quite accurate. At this level of discretization, we see that the conditional multiplicative approximation scheme is within a factor of 2 of the true CP value, but may either underestimate or overestimate depending on which of approximation (A) or (B) from Section~\ref{sec:pathCP} has the stronger effect. This sheds light on a key difference between using MC to estimate path CP as opposed to its alternatives: MC not only gives accurate estimates, but also comes with a \emph{standard error} of that estimate, effectively allowing the user to know whether or not the estimate is a good one. On the other hand, there is no certification for the various other approximations; they simply return point values with no information about how far from the truth they are. This difference is especially crucial given the overarching goal of this exercise, which is to come up with paths that are guaranteed to have a high probability of success and have low cost. The standard error estimates that come from MC can be used as a kind of certificate of accuracy that gives the user confidence in its value, while alternatives come with no such certificate.

\section{Conclusion} 
\label{sec:conclusion}

We have presented a computationally fast method for provably-accurate
pathwise collision probability estimation using variance-reduced Monte Carlo. The
variance-reduction techniques employ a novel planning-specific control
variate and importance distribution. This probability-estimation
technique can be used as a component in a simple meta-algorithm for
chance-constrained motion planning, generating low-cost paths that are
not conservative with respect to a nominal path CP
constraint. Simulation results confirm our theory, and demonstrate
that computation can be done at speeds amenable to real-time
planning.

This works leaves many avenues for further investigation, the foremost
of which is parallelization. As noted earlier, a key feature of MC is
that it is trivially parallelizable (which is not changed by CV or
IS). As most of the computation time 
is spent computing likelihood ratios, which is mostly linear algebra,
our technique is ideally suited for implementation on a GPU. Another
future research direction is to extend this work to more general
controllers and uncertainty models. Heavier-tailed distributions, compared
to the Gaussian model addressed here, would require larger shifts in inflation factor
to affect similar changes in path CP, making a non-conservative CP estimation
procedure all the more important. Monte Carlo itself is extremely
flexible to these parameters, but it remains to be seen if appropriate
control variates or importance distributions can be developed to speed
it up. We note that the meta-algorithm mentioned in this paper is
extremely simple, and can surely be improved upon, although there
was not space in this paper to investigate all such possibilities. One
potential improvement is to incorporate domain knowledge to
differentially inflate the constraints, or to do so in an
iterative or adaptive way, similar in spirit to \cite{OM-BCW:08}. Another improvement 
could be to make bisection search adaptive and to incorporate the
uncertainty in the probability estimates. We also reiterate that
the meta-algorithm can be used with any deterministic planning
algorithm, and thus it is worth exploring which particular algorithms are
best for different planning problems and cost functions. Finally,
although we use our MC method to solve the chance-constrained motion
planning problem, it is in no way tied to that problem, and we plan to
test our method on other problems, such as minimizing CP or optimizing
a objective function that penalizes CP.

\bibliographystyle{plainnat}
{\small
\bibliography{../../../bib/alias,../../../bib/main}
}

\newpage
\appendix
\setcounter{section}{0}
\renewcommand{\thesection}{\Alph{section}}
\section{Problem Formulation}\label{APP:problem}
As briefly discussed in Section~\ref{sec:problem}, we consider the evolution of the robot's path as a discrete approximation of an LQG controller tracking a continuous nominal trajectory. Here we provide the complete characterization of the problem setup.

\subsection{Continuous-time formulation}
We assume the robot's dynamics evolve according to the stochastic linear model:
\begin{equation}
\label{eq:cont}
\dot \x(t) = A_c\x(t) + B_c\u(t)+\v(t), \qquad \quad \y(t) = C_c\x(t) + \w(t),
\end{equation}
where $\x(t) \in \reals^d$ is the state, $\u(t) \in \reals^\ell$ is the control input, $\y(t)$ is the observed output, and $\v \sim \mathcal{N}(\mathbf{0}, V_c)$ and $\w \sim \mathcal{N}(\mathbf{0}, W_c)$
represent Gaussian process and measurement noise, respectively. Let $\Xobs$ be the
obstacle space, so that $\Xfree:= \reals^d\backslash\Xobs$ is the
free space. Let $\Xgoal \subset \Xfree$ and $\xinit \in \Xfree$ be the
goal region and initial state, respectively. 

Let $\xn(t)$ be a nominal solution, i.e., a solution to the deterministic version of the system's equations:
\begin{equation*}
\dot{\x}^{\text{nom}}(t) = A_c \xn(t) + B_c\un (t), \qquad \quad \yn(t)= C_c\xn(t),
\end{equation*}
where $\un(t)$ is the nominal control input, $\yn(t) $ is the nominal measured output,  $\xn_0$ is the (deterministic) initial state, $\xn(T)\in \Xgoal$, and $T$ is the final time. Consider the deviation variables $\dx(t):=\x(t) - \xn(t)$, $\du(t):=\u(t) - \un(t)$, and $\dy(t):=\y(t) - \yn(t)$.
The dynamics of the deviation variables are readily obtained as
\begin{equation*}
\dot \dx(t) = A_c\dx(t) + B_c\du(t)+\v(t), \qquad \quad \dy(t) = C_c\dx(t) + \w(t),
\end{equation*}
where $\dx(0)$ is the initial condition. Consider the quadratic cost functional
\[
J = \mathbb{E}\left[\dx(T)^T F \dx(T)\! +\! \int_0^T \!\!\dx(t)^T Q \dx(t) \!+\! \du(t)^T R \du(t) \right].
\]
The functional $J$ is minimized by the Linear Quadratic Gaussian
(LQG) controller \cite[Chapter 9]{JLS-WHC:08} $\du^{\text{LQG}}(t) =
L(t)\, \widehat \dx(t)$, where $L(t)$ is the linear quadratic regulator (LQR) state feedback
gain and $\widehat \dx(t)$ is the Kalman filter estimate of $\dx(t)$
(we refer the reader to \cite[Chapter 9]{JLS-WHC:08} for further
details on the continuous formulation and provide below detailed equations for the discrete version of the problem).

We are now in a position to state the problem we wish to solve. In
words, we seek to compute  a nominal path of minimum cost subject to
the constraint that the stochastic dynamics driven by a
reference-tracking LQG controller are collision-free with high
probability. More rigorously, for a given nonnegative cost function $c$ (e.g.,
length) acting on paths, the stochastic motion planning problem is defined as
\begin{quote}{\bf Stochastic motion planning (SMP)}:
\begin{equation}
\label{APP_constr}
\begin{split}
\min_{\un(\cdot)} & \quad c(\xn(\cdot)) \\
\text{s.t.} & \quad \prob{\exists t: \x(t)\in \Xobs}  \le \alpha\\
& \quad \u(t) = \un(t) + \du^{\text{LQG}}(t)\\
& \quad \x(0) \sim \mathcal{N}(\xn_0, P_0), \quad \x(T) \in \Xgoal\\
& \quad \text{Equation \eqref{eq:cont}}   
\end{split}
\end{equation}
\end{quote}

\subsection{Discrete Approximation}

To solve the SMP problem, one is required to compute path collision probabilities $\prob{\exists t: \x(t)\in \Xobs}$. To make the problem tractable, we consider a discrete formulation whereby the dynamics are given by
\begin{equation}\label{eqn:APP_DLQG}
\x_{t+1} = A\x_t + B\u_t+\v_t, \quad \v_t \sim \mathcal{N}(\mathbf{0},
V), \quad \y_t = C\x_t + \w_t, \quad \w_t \sim \mathcal{N}(\mathbf{0}, W),
\end{equation}
for a given fixed timestep $\dt$. In equation \eqref{eqn:APP_DLQG}, we used the definitions
\begin{equation*}
A := e^{A_c\dt}, \quad B := \left(\int_0^\dt e^{A_cs} ds\right)B_c,
\quad C := C_c, \quad V := \int_0^\dt e^{A_cs}V_c e^{A^T_cs} ds, \quad W := \frac{1}{\dt}W_c.
\end{equation*}
The deviation variables are then defined as $\dx_t:=\x_t - \xn_t$, $\du_t:=\u_t - \un_t$, and $\dy_t:=\y_t - \yn_t$, for $t = 0,\dots,T$. The cost function for the discrete LQG controller is given by 
\[
J = \mathbb{E}\left[\dx_T^T F \dx_T + \sum_{t=0}^{T-1} \dx_t^T Q \dx_t + \du_t^T R \du_t \right],
\]
and the LQG controller minimizing $J$ is given in \cite{AJS-IRP:08} by
$\du^{\text{LQG}}_t = L_t \, \widehat \dx_t$, where the feedback gain is
\begin{align*}
L_t &= -(R + B^T S_{t+1} B)^{-1} B^T S_{t+1} A,\\
S_t &= Q + A^T \left(S_{t+1} - S_{t+1} B (R + B^T S_{t+1} B)^{-1} B^T
  S_{t+1}\right) A, \qquad S_T = F,
\end{align*}
and the Kalman estimate dynamics are 
\[
\widehat \dx_{t+1} = A \widehat \dx_t + B L_t \widehat \dx_t + K_t (\dy - C \widehat \dx_t),
\]
\begin{equation*}
K_t = AP_t C^T (W + CP_t C)^{-1}, \qquad P_{t+1} = V + A \left(P_t - P_t C^T (W + CP_t C)^{-1} C P_t\right) A^T,
\end{equation*}
with $P_0$ denoting the covariance of the initial state $\x_0$ (and hence of $\dx_0$). The combined  system evolves according to 
\begin{equation}\label{eqn:DLQGcomb}
\begin{bmatrix} \dx_{t+1}\\ \widehat \dx_{t+1} \end{bmatrix} =
\begin{bmatrix} A & BL_t\\ K_tC & A + BL_t - K_tC\end{bmatrix}
\begin{bmatrix} \dx_t\\ \widehat \dx_t \end{bmatrix}
+ \begin{bmatrix} \v_t\\ K_t\w_t \end{bmatrix} = M_t \begin{bmatrix} \dx_t\\ \widehat \dx_t \end{bmatrix} + \n_t,
\end{equation}
where $\n_t \sim \mathcal{N}\left(\mathbf{0}, N_t = \begin{bmatrix} V & 0\\ 0 & K_tWK_t^T\end{bmatrix}\right)$.
Equation \eqref{eqn:DLQGcomb}, in addition to providing a formula for simulating
state trajectories, also represents a means for tracking the state uncertainty at
each time step. Given that $\begin{bmatrix} \dx_{t}\\ \widehat \dx_{t} \end{bmatrix} \sim \mathcal{N}\left(\mu_t, \Sigma_t\right)$,
we have that
\begin{equation}\label{eqn:APPuncertaintyevol}
\begin{bmatrix} \dx_{t+1}\\ \widehat \dx_{t+1} \end{bmatrix} \sim \mathcal{N}\left(\mu_{t+1} = M_t\mu_t, \Sigma_{t+1} = M_t\Sigma_tM_t^T + N_t\right).
\end{equation}
Using this recursion we may compute the marginal distributions of each waypoint along
the full LQG-controlled trajectory, starting from $\mu_0 = \mathbf{0}$
and $\Sigma_0 = \begin{bmatrix} P_0 & 0\\ 0 & 0\end{bmatrix}$. Letting
$\overline{\x_0,\dots,\x_T}$ denote the continuous curve traced by the
robot's random trajectory (which connects the points $\x_0,\dots,\x_T$), the problem we wish to solve then becomes:
\begin{quote}{\bf Discretized SMP}:
\begin{equation}
\label{APP_Dconstr}
\begin{split}
\min_{\un(\cdot)} & \quad c(\xn(\cdot)) \\
\text{s.t.} & \quad \prob{\overline{\x_0,\dots,\x_T} \cap \Xobs \neq \varnothing}  \le \alpha\\
& \quad \u_t = \un_t + \du^{\text{LQG}}_t\\
& \quad \x_0 \sim \mathcal{N}(\xn_0, P_0), \quad \x_T \in \Xgoal\\
& \quad \text{Equation  \eqref{eqn:APP_DLQG}}   
\end{split}
\end{equation}
\end{quote}

\end{document}